\documentclass[11pt]{article}
\usepackage{coling2020}
\usepackage{times}
\usepackage{url}
\usepackage{latexsym}

\usepackage{microtype}
\usepackage{booktabs}
\usepackage{makecell}
\usepackage{tabularx}
\usepackage{colortbl}
\usepackage{amsthm}
\usepackage{amssymb}
\usepackage{amsmath}
\usepackage{amsfonts}
\usepackage{bm}
\usepackage{arydshln}
\usepackage{comment}
\usepackage{xspace}
\usepackage{mathptmx}
\usepackage{enumitem}
\usepackage{titlesec}

\titlespacing{\paragraph}{0pt}{.5pc}{1pc}

\usepackage{subcaption}
\usepackage{graphicx}
\usepackage{multirow}

\usepackage[textsize=footnotesize]{todonotes}
\definecolor{Gray}{gray}{0.95}
\newcommand{\cntodo}[1]{\todo[color=green!20]{#1}}
\newcommand{\altodo}[1]{\todo[color=yellow!20]{#1}}
\newcommand{\jttodo}[1]{\todo[color=blue!20]{#1}}
\newcommand{\corpus}{\texttt{GAQCorpus}\xspace} 
\newcommand{\corpuslong}{Grammarly Argument Quality Corpus\xspace} 

\colingfinalcopy 

\title{\emph{Rhetoric, Logic, and Dialectic}:\\ Advancing Theory-based Argument Quality Assessment\\ in Natural Language Processing}

\author{Anne Lauscher\textsuperscript{1}, Lily Ng\textsuperscript{2}, Courtney Napoles\textsuperscript{2},  Joel Tetreault\textsuperscript{3} \vspace{0.3em} \\
  \textsuperscript{1}Data and Web Science Group, University of Mannheim, Germany \\
  \textsuperscript{2}Grammarly 
  \textsuperscript{3}Dataminr, Inc.
  \\ 
  \textsuperscript{1}{\tt anne@informatik.uni-mannheim.de}, \\
  \textsuperscript{2}{\tt first.last@grammarly.com}, 
  \textsuperscript{3}{\tt jtetreault@dataminr.com}
}

\date{}

\begin{document}
\maketitle
\begin{abstract}
Though preceding work in computational argument quality (AQ) mostly focuses on assessing overall AQ, researchers agree that writers would benefit from feedback targeting individual dimensions of argumentation theory. 
However, a large-scale theory-based corpus and corresponding computational models are missing. 
We fill this gap by
conducting an extensive analysis covering three diverse domains of online argumentative writing and
presenting \corpus: the first large-scale English multi-domain (community Q\&A forums, debate forums, review forums) corpus annotated with theory-based AQ scores. We then propose the first computational approaches to theory-based assessment, which can serve as strong baselines for future work. We demonstrate the feasibility of large-scale AQ annotation, show that exploiting relations 
between dimensions yields performance improvements, and explore the synergies between theory-based prediction and practical AQ assessment.

\end{abstract}

\section{Introduction}
Providing relevant and sufficient justifications for a claim and using clear language to express reasoning are important features of everyday writing.  
These are components of \emph{Argument Quality (AQ)}, which has been studied in many domains, such as student essays~\cite{wachsmuth-etal-2016-using}, news editorials~\cite{el-baff-etal-2018-challenge}, and debate forums~\cite{lukin-etal-2017-argument}.

Preceding work in natural language processing (NLP) and computational linguistics (CL) has mostly focused on practical AQ assessment\footnote{We adopt the terminology of \newcite{wachsmuth-etal-2017-argumentation} who refer to task-driven approaches, which often also focus on the \emph{relative} assessment of AQ, as ``practical''.}, 
considering either the \emph{overall quality} of arguments \cite[inter alia]{toledo2019automatic,gretz2019large} or a single specific conceptualization of AQ, e.g.,  \textit{argument strength}~\cite{persing2015modeling}, \textit{convincingness}~\cite{habernal-gurevych-2016-argument}, and \textit{relevance}~\cite{wachsmuth-etal-2017-pagerank}. However, \newcite{gretz2019large} note the need to predict quality in terms of fine-grained aspects. 
Fine-grained prediction enables a deeper understanding of argumentation and offers specific feedback to authors aiming to improve their argumentative writing skills. For instance, authors might want to know whether their premises are \emph{sufficient} with regard to their claim(s) or whether their language is \emph{appropriate}.
\newcite{wachsmuth-etal-2017-computational} surveyed and synthesized theory-based dimensions of AQ into a taxonomy of three main dimensions: Cogency (Logic), Effectiveness (Rhetoric), and Reasonableness (Dialectic). Their initial annotation study showed that assessing these dimensions is challenging, even for experts, but 
that crowd workers can handle the task comparably well if the guidelines and task are simplified.

Given the feasibility of annotation and the recognized need for fine-grained dimensions in AQ assessment, it is surprising that 
no further efforts in NLP and CL have been made. There is no large scale annotated corpus and, consequently, no computational model. 
In this work, we aim to fill this research gap by conducting an in-depth analysis of theory-based AQ assessment covering overall AQ and the three dimensions (logic, rhetoric, and dialectic) of the Wachsmuth et al. taxonomy, and three diverse domains of online argumentative writing (Q\&A forums, debate forums, and review forums).

Drawing on existing AQ theories, we address five research questions
(\textbf{RQs}) to inform and fuel future AQ annotation studies and computational AQ research:

\textbf{RQ1:} \textit{Can we develop a large-scale theory-based AQ corpus?}
We conduct an extensive annotation study with trained linguists and crowd workers on $5,295$ arguments from three domains to create the \corpuslong (\corpus), the first large-scale multi-domain English corpus annotated with theory-based AQ scores. 

\textbf{RQ2}: \textit{Are we able to develop computational models that can do theory-based AQ assessment in varying domains?}  Based on \corpus, we
are the first to propose computational approaches to theory-based AQ assessment and show that it is possible to develop models for this task. Our models can serve as strong baselines for future research and enable the field to investigate follow-up research questions. 

\textbf{RQ3}: \emph{Can the interrelations between the different AQ dimensions be exploited in a computational setup?} 
Inspired by the hierarchical structure of the taxonomy,
we explore whether the relationships between dimensions can be computationally exploited.
In addition to simple single-task learning approaches, we study the effect of jointly predicting AQ dimensions in two variants (\emph{flat} vs. \emph{hierarchical}) and find
that combining the training signals of all four aspects benefits theory-based AQ assessment. 

\textbf{RQ4}: \emph{Does the corpus support training a single unified model for multi-domain evaluation?}
When enough data from a single domain is available, training on in-domain data is typically preferred over multi-domain. However, larger amounts of data are especially useful for complex model architectures currently prominent in NLP (e.g., BERT~\cite{devlin2018bert}, GPT2~\cite{radfordlanguage}).
We study these two mutually opposing effects on \corpus and show that our corpus supports training a single unified model across all three domains,
with improved performances in individual domains. 

\textbf{RQ5}: \emph{Can we empirically substantiate the idea that theory-based and practical AQ assessment can learn from each other?} 
\newcite{wachsmuth-etal-2017-argumentation} suggest that both the practical and the theory-based 
views can learn from each other, but so far, this has been only tested manually. Employing our models, we go one step further and conduct a bi-directional experiment employing a practical AQ corpus. We demonstrate two concrete ways how theory-based and practical AQ research can profit from their combination.

\paragraph{Structure.} After discussing related work in \S \ref{sec:related-work}, we describe our annotation study and resulting corpus (\S \ref{sec:annotation-study}). \S \ref{sec:methodology} describes the computational approaches which we employ in the experiments (\S \ref{sec:experiments}). Last, we conclude our work and give potential directions for future work (\S \ref{sec:conclusion}).


\section{Related Work}
\label{sec:related-work}
Earlier work in computational AQ assessment can be divided into practical and theory-based approaches.


\paragraph{Practical approaches.} Recently, the field of computational AQ research has been mostly driven by practical approaches 
that each target an individual domain. 
Accordingly, past approaches tackle either overall quality \cite{toledo2019automatic} or specific subqualities of argumentation, such as convincingness \cite{habernal-gurevych-2016-argument} and relevance \cite{wachsmuth-etal-2017-pagerank}. 
The popularity of practical approaches can partly be attributed to the relative simplicity of crowd-sourcing annotations. 

Much prior work has focused on aspects of student essays, including essay clarity \cite{persing-ng-2013-clarity}, organization \cite{Persingorganization}, prompt adherence \cite{persing-ng-2014-prompt}, and argument strength \cite{persing2015modeling}. 
Later, \newcite{wachsmuth-etal-2016-using} present an approach driven by detecting argumentative units, thereby demonstrating the usefulness of argument mining techniques to the problem. Similarly,
\newcite{stab2016recognizing} predict the absence of opposing arguments and \newcite{stab-gurevych-2017-recognizing} predict insufficient premise support in arguments.
Another well-studied domain is web debates. \newcite{wachsmuth-etal-2017-pagerank} adapt PageRank 
 to identify argument relevance. Pairwise comparison of the convincingness of debate arguments has been conducted \cite{habernal-gurevych-2016-argument}. \newcite{Persing:2017:WCY:3171837.3171856} additionally predict why an argument receives a low persuasive power score. By explaining flaws in argumentation, they highlight the importance of explainability and specific author feedback.

Other approaches take into account properties of the source, i.e., the author \cite{cardieuser} or the audience \cite{el-baff-etal-2018-challenge,durmus-cardie-2018-exploring}.
In contrast, we assume that a system may not have much knowledge about the authors or audience and thus our models operate solely on the text. 
\newcite{toledo2019automatic} and \newcite{gretz2019large} present large corpora of crowd-sourced arguments and their quality. These corpora cover a variety of topics, but only within single domains. The authors emphasize that research on theory-based approaches could further advance the field of computational AQ. 

\paragraph{Theory-based approaches.} Rooted in classic argumentation theory, the works can according to \newcite{wachsmuth-etal-2017-computational}, be categorized based on whether they related to the \emph{logical} \cite{johnson2006logical,hamblin1970fallacies}, \emph{rhetorical} \cite{kennedy2007aristotle}, or \emph{dialectical} \cite{perelman1971new,van2004systematic} properties of an argument.

\newcite{wachsmuth-etal-2017-computational} were the first to survey and highlight the importance of the theory-based 
approach to computational AQ and synthesized the argumentation-theoretic literature into a taxonomy. 
\newcite{wachsmuth-etal-2017-argumentation} conducted a study in which crowd workers annotated 304 arguments for all 15 quality dimensions following \newcite{wachsmuth-etal-2017-computational},
and demonstrated that the theory-based and practical AQ assessment match to a large extent and that the two views can learn from each other, for instance, when it comes to more practical annotation processes for theory-based AQ annotations. 

However, until now, no further research on computational theory-based AQ assessment in NLP has been conducted, no larger-scale annotated corpus has been presented, and thus
no computational model that would allow further investigation into the concrete synergies between the two perspectives exists.

\section{Annotation Study}
\label{sec:annotation-study}
\newcite{wachsmuth-etal-2017-argumentation} suggest that large-scale annotation of theory-based AQ dimensions is possible. We test this finding and take it one step further by asking whether we can develop a large-scale theory-based AQ corpus (\textbf{RQ1}). This section presents \emph{\corpus}, the result of the first study annotating theory-based dimensions, including $5,285$ arguments from three diverse domains of real-world argumentative writing.

\subsection{Annotation Scheme}\label{sec:annotation-scheme}
%
%

\begin{figure}[t!]
\centering
\begin{minipage}[t]{0.54\linewidth}\strut\vspace*{-\baselineskip}\newline
    \centering
    \includegraphics[width=1.0\linewidth,trim=0.2cm 0cm 0.2cm 0cm]{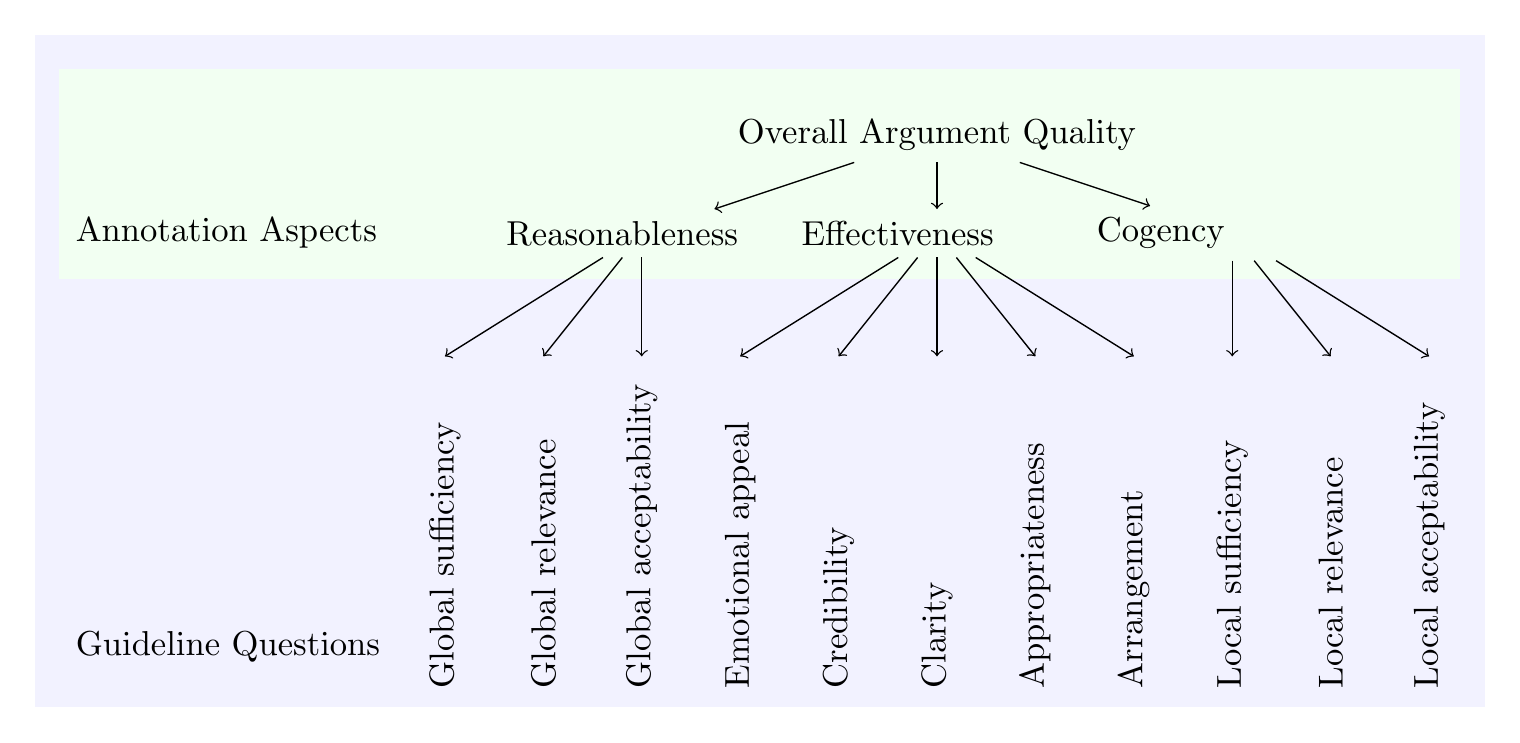}
    \caption{Taxonomy of theory-based AQ \cite{wachsmuth-etal-2017-computational}. Questions related to each aspect guided annotators in assessing higher level dimensions.}
    \label{fig:tax}
    \end{minipage}
    \hfill
    \centering
\begin{minipage}[t]{0.44\linewidth}\strut\vspace*{-\baselineskip}\newline
    \centering
    \includegraphics[width=1.0\linewidth,trim=0.25cm 0.0cm 1.5cm 0cm]{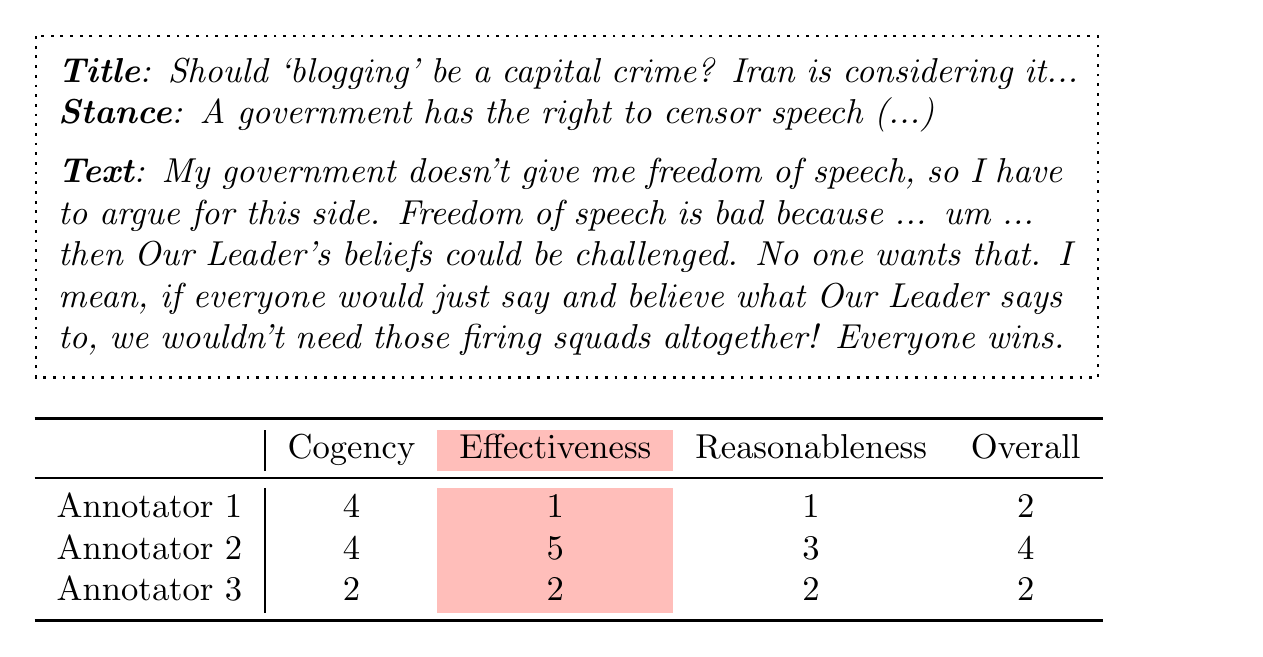}
    \caption{Example text from our annotation pilot. Linguistic expert annotators highly disagree on scoring the effectiveness dimension.}
    \label{fig:ex1}
\end{minipage}
\end{figure}

Our annotation scheme is based on the \newcite{wachsmuth-etal-2017-argumentation} taxonomy of argumentation quality  depicted in Figure~\ref{fig:tax}. 
It defines \textbf{overall AQ} as being composed of three sub-dimensions (Cogency, Effectiveness, Reasonableness), each of which is in turn composed of several quality-related aspects: 
\begin{itemize}[nosep]
\item\textbf{Cogency} relates to the logical aspects of AQ. High cogency indicates that an argument's premises are acceptable as well as relevant and sufficient with regard to the argument's conclusion.
\item\textbf{Effectiveness} reflects the persuasive power of how an argument is stated. Important aspects of an effective argument include its arrangement, clarity, appropriateness in a given context, emotional appeal, and author's credibility.
\item\textbf{Reasonableness} indicates the quality of an argument in the context of a debate, i.e., its relevance, its acceptability and the way it is stated as a whole, and its sufficiency toward the resolution of the issue.
\end{itemize}
\noindent Starting from the guidelines of \newcite{wachsmuth-etal-2017-computational}, we developed our annotation guidelines through a series of pilot studies with four \textit{expert annotators} who are all fluent or native English speakers with advanced degrees in linguistics.
\newcite{wachsmuth-etal-2017-argumentation} recommend simplifying the task and guidelines, and based on the findings of our pilots, we made the following modifications under consultation with our experts:
Since the annotators noted difficulties distinguishing between the 15 fine-grained aspects, we collapse the scheme to Overall AQ and the three higher level dimensions and represent the finer-grained sub-dimensions as questions to guide the annotators' judgments. 
We additionally use a five-point scale (very low, low, medium, high, very high, plus ``cannot judge''), which simplifies the task according to feedback from our expert annotators and previous findings~\cite{cox1980}. We experimented with both the five-point and the original three-point rating scale (low, medium, high) used by \newcite{wachsmuth-etal-2017-computational}, and found that switching the scales did not negatively affect inter-annotator agreement.

\newcite{ng2020aqcorpus} describe the annotation design and guidelines in more detail.


\subsection{Data} 
\begin{figure}[t!]
\centering
\begin{minipage}[t]{1.0\linewidth}
    \centering
    \includegraphics[width=1.0\linewidth,trim=0.2cm 0.2cm 0.2cm 0.1cm]{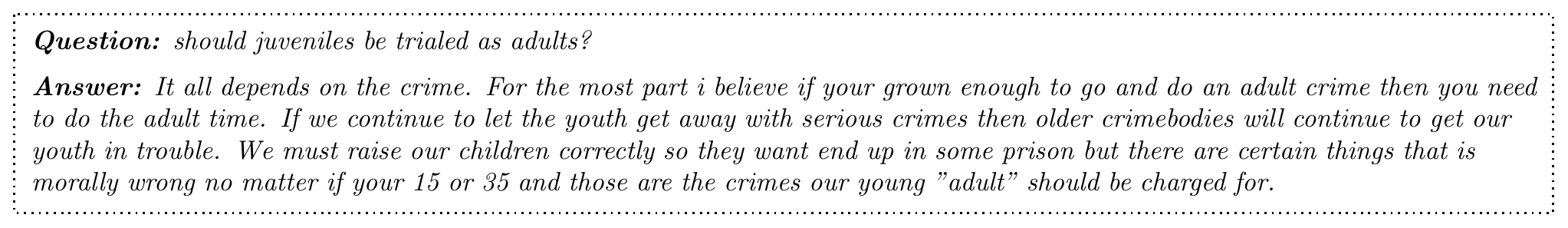}
    \subcaption{Community Q\&A Forums.}
    \label{fig:ex_deb}
    \end{minipage}
\begin{minipage}[t]{1.0\linewidth}

    \centering
    \includegraphics[width=1.0\linewidth,trim=0.2cm 0.2cm 0.2cm 0.1cm]{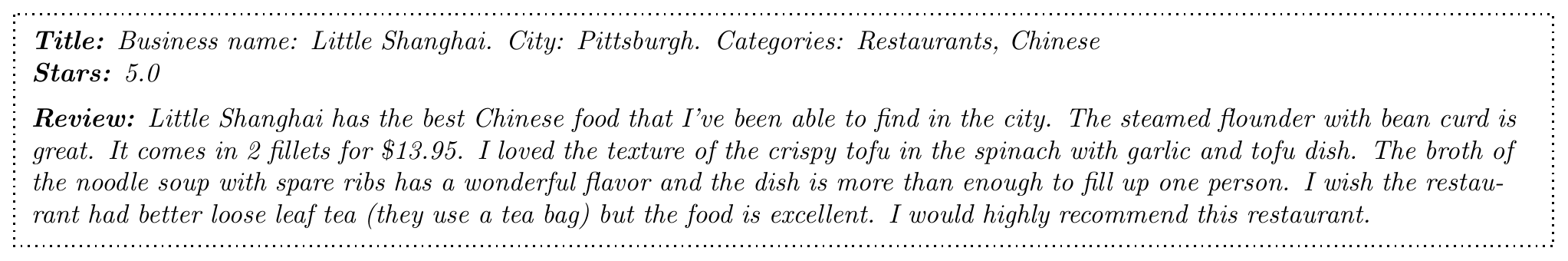}
    \subcaption{Review Forums.}
    \label{fig:ex_qa}
\end{minipage}
\caption{Example texts and quality trends provided by our linguistic experts.}
\label{fig:domain_ex}
\end{figure}

We investigate different domains to obtain a deeper understanding of real-world AQ and the feasibility of the annotation scheme in different settings. We include 
three domains in our study: Community questions and answers forum posts (\textit{CQA}), debate forum posts (\textit{Debates}), and business review forum posts (\textit{Reviews}). 
Figures~\ref{fig:ex1} and~\ref{fig:domain_ex} display an example text for each domain. 


\paragraph{CQA.} We include 2,088 arguments from Yahoo! Answers,\footnote{\url{https://answers.yahoo.com/}} a community questions and answers forum where users ask questions and answer questions posted by others. Not enforcing strict debating rules or topics, the argumentative posts are diverse and therefore interesting for our study. 
While not a dedicated debate forum, we found that some categories contain a relatively high proportion of argumentative posts, like \emph{Politics \& Government $\rightarrow$ Law \& Ethics}, from which we select posts.  
We only include posts marked as best answer for a question and exclude posts containing uniform resource identifiers or media content. 
From these, we select posts that were labeled as argumentative by a majority of 10 raters in a secondary experiment (see Appendix).


\paragraph{Debates.} To reflect online debate forums-style argumentation, we include 1,337 arguments from Change My View (CMV) \cite{tan+etal:16a} and 766 from the Internet Argument corpus V2 (IAC) \cite{abbott-etal-2016-internet} resulting in 2,103 arguments in total. 
CMV is an internet forum in which users post their opinion and ask others to challenge their beliefs on the topic. 
The IAC is composed of posts retrieved from three online debate forums, and in this study we include only arguments from the ConvinceMe subset.
We try to restrict the sample to instances that do not require much background knowledge or thread-level context. From CMV, we include original posts only and for ConvinceMe, we include the first post reacting to the topic.  From CMV we also exclude posts tagged \texttt{[MOD]}, which indicate moderator posts. 

\paragraph{Reviews.}
 Yelp is an online platform where users publish business reviews and rate their experience from 1 (poor) to 5 (excellent) stars. From the Yelp-Challenge-Dataset\footnote{\url{https://www.yelp.com/dataset}}, we sampled 1,104 arguments reviewing restaurants. While the review texts often do not appear as ``classic" arguments, i.e., with a dedicated claim and premises supporting this claim, 
the texts can indeed be considered argumentative \cite{wachsmuth2014review,wachsmuth2015sentiment}; The star rating corresponds to a claim a user is making about the business and the review text is intended to support this claim.  

\smallskip\noindent 
For Debate and Review posts, we include the star rating and stance (if provided) with the text.  Across all domains, we filter for posts with text length between $70$ and $200$ words. 
%
%
%
%
%
To ensure high quality annotations, we first ran $13$ pilot studies in two flavors: (1) with three of the linguistic expert annotators (\S\ref{sec:annotation-scheme}), 
and (2) with a crowd-sourced workforce of $24$ contributors from Appen.\footnote{Formerly \textit{Figure Eight}, \url{https://www.appen.com/}} 
Both groups  used the same annotation guidelines and interface, which we iteratively improved based on feedback from each pilot. 
Table~\ref{tab:num_annotators} shows the number of judgments per instance per domain as well as the number of instances that were annotated by each group. For each domain, up to 538 arguments were annotated by both experts and crowd workers. 
%
%
%
\begin{table}[!t]
\centering
\parbox[t]{.49\linewidth}{
\strut\vspace*{-\baselineskip}\newline
\centering
{\small
%
%
\begin{tabular}{lrrrrrr}\toprule
&Crowd &\multicolumn{3}{c}{Experts} &Overlap \\
\cmidrule(lr){2-2} \cmidrule(lr){3-5} \cmidrule(lr){6-6}
\# Annotators &10 &1 &2 &3 & 11--13\\
\midrule
CQA &1,334 &626 &-- & 625 &500 \\
Debates &1,438 &600 &-- &600 &538 \\
Reviews &600 &200 &400 &-- &100 \\
\bottomrule
\end{tabular}
\caption{Number of annotators per instance and total instances annotated by Experts and the Crowd, and the number of overlapping instances by domain.}\label{tab:num_annotators}
}}
\hfill
\parbox[t]{.46\linewidth}{
\strut\vspace*{-\baselineskip}\newline
\centering
\small
\begin{tabular}{lrrrrr}\toprule
Domain & Total & Train & Dev. & Test 
\\\midrule
CQA  &2,085 &1,109 &476 &500 
\\
Debates &2,100 &1,093 &469&538 
\\
Reviews &1,100 &700 &300 &100 
\\
\midrule
All &5,285 &2,902 &1,245 &1,138 
\\
\bottomrule
\end{tabular}
\caption{Number of instances in the train, development, and test sets of \corpus.}\label{tbl:corpus}

}
\end{table}
%
%

We provide and use a standard split for each domain, which is composed as follows: The training and development sets consist of the instances which were \emph{either} annotated by our linguistic experts or the crowd workers.  
In contrast, the test sets encompass only instances  scored by \emph{both} experts and the crowd. For each instance and group, we obtain a single score by averaging the annotators' votes. In addition to the group-specific annotations (\emph{expert} and \emph{crowd}), we also compute a \emph{mix} score which is the average of the two group-specific scores. This way, we train on a mix of expert and crowd annotations (where the dominant portion comes from the crowd) and test on overlapping instances, enabling us to compare  model performance to both expert and crowd ratings on a static set of instances. 
The numbers of instances in each portion of \corpus are given in Table~\ref{tbl:corpus}.

\subsection{Data Analysis}
\setlength{\tabcolsep}{3pt}
\begin{table*}[!t]
\parbox[t]{.48\linewidth}{
\strut\vspace*{-\baselineskip}\newline
\centering
{\small
\begin{tabularx}{\linewidth}{ l c c c c}
\toprule
& \multicolumn{1}{c}{Cogency} & \multicolumn{1}{c}{Effectiveness} & \multicolumn{1}{c}{Reasonableness} & \multicolumn{1}{c}{Overall} \\
\midrule
Ours & \textbf{0.46} 
& \textbf{0.48} 
& \textbf{0.48} 
& \textbf{0.55} \\
TvsP & 0.27 
& 0.38 
& 0.13 
& 0.43 
\\
\bottomrule
\end{tabularx}
}
\caption{Agreement between expert annotations from \newcite{wachsmuth-etal-2017-computational} and crowd-sourced annotations from two sources: \corpus (Ours) and \textsc{TvsP} on $200$ randomly sampled instances. 
}
\label{tbl:dagstuhl}
}
\hfill
\parbox[t]{.48\linewidth}{
\strut\vspace*{-\baselineskip}\newline
\centering
\small
\begin{tabular}{lcccc}\toprule
& Cogency &Effectiveness &Reasonableness &Overall \\\midrule
CQA  &0.42 &0.52 &0.52 &0.53 \\
Debates\hspace{-1em} &0.14 &0.11 &0.21 &0.19 \\
Reviews\hspace{-1em} &0.32 &0.32 &0.31 &0.33 \\
\bottomrule
\end{tabular}
\caption{IAA between the Expert and Crowd scores for Cogency, Effectiveness, Reasonableness, and Overall AQ.}\label{tab:agreement_expert_crowd}
}

\end{table*}
%
%
%


\paragraph{Inter-annotator Agreements (IAA).\label{iaa}} To assess the quality of our crowd-sourced annotations and 
%
%
our simplified guidelines, we employ the  Dagstuhl-ArgQuality-Corpus-V2 (\textsc{DS}, originally from UKPConvArgRank \cite{habernal-gurevych-2016-argument}) 
and conduct a comparative study against the crowd-sourced \newcite{wachsmuth-etal-2017-argumentation} annotations ({\sc TvsP}).
We take ``gold'' ratings from the original, author-produced annotations of \newcite{wachsmuth-etal-2017-computational}. \textsc{DS} was presented in combination with the taxonomy of theory-based AQ described above and consists of $304$ web debate arguments annotated with all $15$ AQ aspects. 
We randomly sample $200$ arguments and crowd-source annotations on Appen with our revised guidelines.\footnote{Here, we stuck to the original 3-point scale to match the original expert annotations we compare with.}
For each instance and AQ dimension, we collect 10 votes and average them to obtain the group vote (Mean). 
We measure IAA between the group vote and the \textsc{DS} expert vote with Krippendorff's $\alpha$ \cite{krippendorff2011computing}. 
The results are depicted in Table~\ref{tbl:dagstuhl}. 
The agreement does not exceed 0.55, which is not surprising for a task of this subjectivity, and generally, the agreement scores of our crowd ratings 
 surpass the agreement scores reported by {\sc TvsP}. we therefore conclude that our guidelines and user interface support the task and confirm the suitability of our crowd annotators. 

Next we consider the 
agreement between experts and crowd workers on the overlapping portions of \corpus using the mean scores (Table~\ref{tab:agreement_expert_crowd}). For debate forums, Krippendorff's $\alpha$ is up to $0.21$, while for the Q\&A forums, the agreement is higher -- up to $0.53$. These results suggest that the difficulty of the task is highly dependent on the domain. 
%
\paragraph{Analysis of Disagreements.}

We noticed disagreements among the annotators along all stages of the annotation process, especially for arguments which were of sarcastic or ironic nature or included rhetorical questions. Consider the argument given in Figure~\ref{fig:ex1} as an example. 

This example on the topic of \textit{freedom of speech} seems to support the stance that a government has the right to censor speech. 
However, several linguistic cues indicate that the argument might be ironic: (a) Punctuation: Ellipsis indicates thinking/searching for justifications; similarly, (b) the filler \textit{um}; (c) Capitalization: The noun phrase \textit{Our Leader} is capitalized, indicating hyperbolic apotheosis; and finally, (d) the phrase \textit{(...) so I have to argue for this side.} acts like an apologia, which is put in front of the actual argument. In discussion with our expert annotators, it became clear that Annotators~$1$ and ~$2$ based their judgments on an interpretation of this text that related to the estimated degree of irony in the post. While Annotator~$1$ did not perceive irony and judged the argument as \textit{very weak} in \textit{Effectiveness}, Annotator~$2$ considered it to be highly effective as in their view, the irony positively underlined the perceived stance. Annotator~$3$ gave medium scores across the board but was leaning more towards Annotator~$2$'s opinion. Such disagreements were regularly discussed and usually revealed that multiple opinions may exist according to how the texts were interpreted, which highlights the high subjectivity of the task. 

Disagreements can also be observed across different domains. Debates and CQA are dialectic by nature, but
original posts (or top answers in the case of CQA) are relatively straightforward to assess in isolation. In contrast, business reviews are monologues and cite experiences as justifications for a claim, e.g., \textit{My meal was delicious}. Given that experience is subjective, evaluating reviews presents unique challenges.

\begin{figure}
     \centering
     \begin{subfigure}[t]{0.49\textwidth}
         \centering
         \includegraphics[width=1.0\linewidth,trim=0.0cm 0cm 1.5cm 0cm]{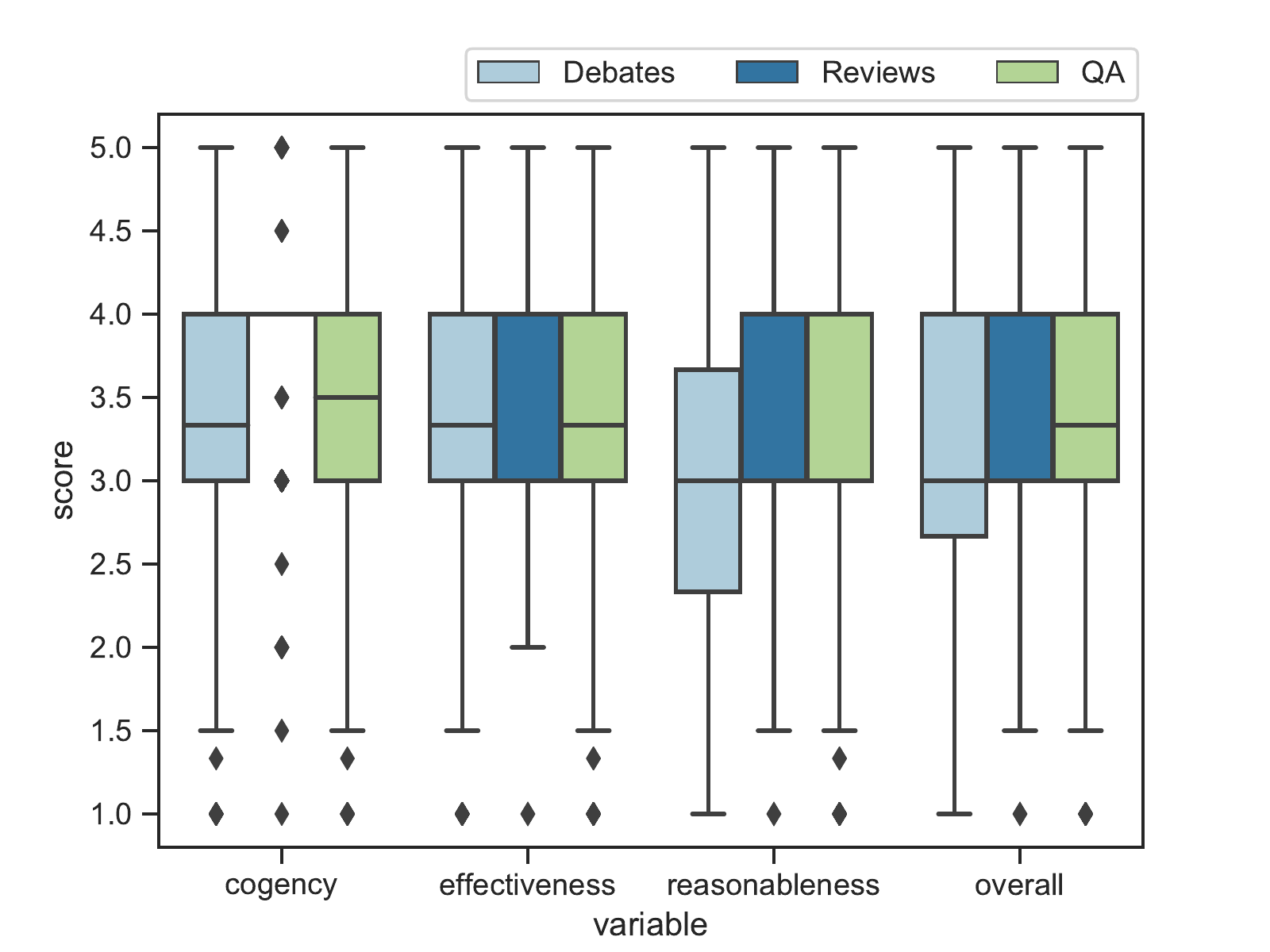}
         \caption{Expert annotations.}
         \label{fig:domains_experts}
     \end{subfigure}
     \begin{subfigure}[t]{0.49\textwidth}
         \centering
         \includegraphics[width=1.0\linewidth,trim=0.0cm 0.0cm 1.5cm 0cm]{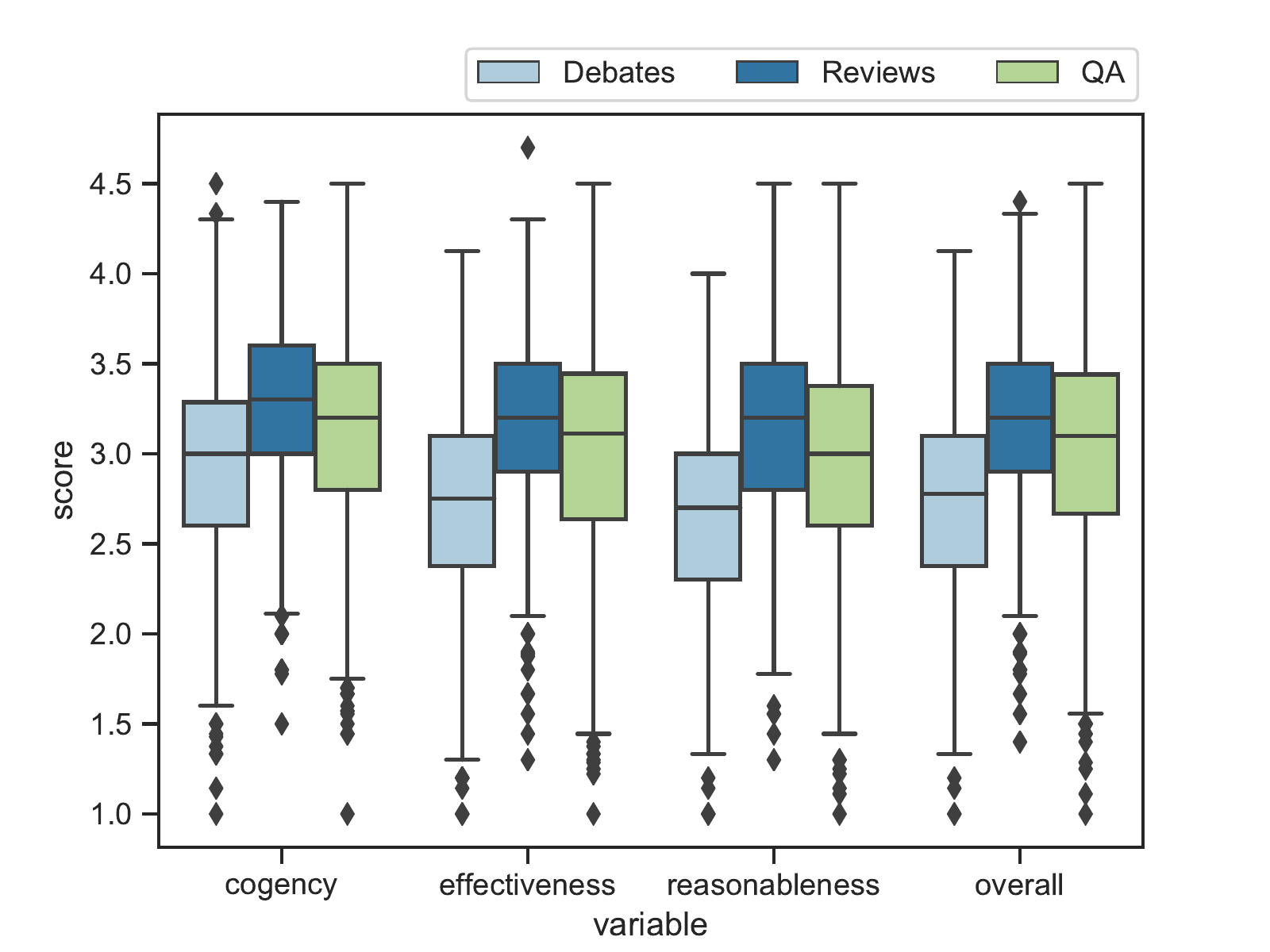}
         \caption{Crowd annotations.}
         \label{fig:domains_crowd}
     \end{subfigure}
        \caption{Score distributions of our variables (cogency, effectiveness, reasonableness, overall argument quality) by domain (debate forums, review forums, Q\&A forums) for expert and crowd annotators.}
        \label{fig:domains}
\end{figure}
%
%
%
The final distributions of the mean scores per variable across the different domains in \corpus are depicted in Figures \ref{fig:domains_experts} and \ref{fig:domains_crowd}, 
respectively. The interquartile range of the expert scores is generally higher than for the crowd annotations, which might indicate that experts are more specific when scoring examples. This is also reflected in the medians across the different variables: While the crowd exhibits a tendency to score variables equally, more differentiation can be seen in the experts' annotations. 

%
%
%
%
%


\section{Models}
\label{sec:methodology}
Having developed \corpus to enable computational AQ assessment (\textbf{RQ1)}, we address the remaining research questions by experimenting with AQ models. To determine whether we can  develop a computational theory-based AQ model (\textbf{RQ2)}, we employ a naive length baseline, three different Support Vector Regression (SVR)  models, and a BERT-based \cite{devlin2018bert} model.
We next investigate whether the interrelations between AQ dimensions can be exploited in a computational setup (\textbf{RQ3}), employing two multi-task BERT-based models.
For the BERT-based models, we transform each argument into a ``BERT-compatible'' format, i.e., into a sequence of WordPiece \cite{wu2016google} tokens and prepend the sequence with \textsc{BERT}'s start token (\texttt{[CLS]}). The pooled hidden representation of the latter corresponds to the aggregated document representation. 
The specific details of each model are described below.


\paragraph{Argument Length (\textsc{Arg length}).} To estimate the task difficulty and to measure a potential length bias, our naive baseline is the correlation between the argument's character length and quality scores.

\paragraph{SVR with Lexical Features (\textsc{SVR$_{tfidf}$}).} We run a simple SVR with tf--idf feature vectors. 

\paragraph{SVR with Semantic Features (\textsc{SVR$_{embd}$}).} We represent each argument as the average of the fastText~\cite{Bojanowski:2017tacl} embedding\footnote{\url{https://dl.fbaipublicfiles.com/fasttext/vectors-english/wiki-news-300d-1M-subword.vec.zip}} representation of each word in the argument.

\paragraph{Feature-rich SVR (\textsc{Wachsmuth$_{CFS}$}).}
%
 We reimplement 
 the approach of \newcite{wachsmuth-etal-2016-using}, who employ standard features (token n-grams, part-of-speech tags, etc.) and higher-level features (sentiment flows, argumentative discourse units etc.). We run correlation-based feature selection on the training set and include only the most predictive features.

\paragraph{Single Task Learning Setting (\textsc{BERT ST}).} For each AQ dimension, 
we train an individual regressor.  
  Our AQ predictor is a simple linear regression layer in which we feed the pooled document representation. 
%
The loss $L_{\mathit{t}}$ is then simply the mean squared error (MSE) over $k$ instances in the training batch.
%
\paragraph{Flat Multi-Task Learning Setting (\textsc{BERT MT$_{flat}$}).} 
We explore whether a joint training setup would improve the individual score predictions. 
For each quality dimension, we employ an individual prediction layer as described above and compute an individual task loss. 
We then define the total training loss 
as the sum of the task losses. 
%
%

\paragraph{Hierarchical Multi-Task Learning Setting (\textsc{BERT MT$_{hier}$}).} 
We propose a hierarchical multi-task learning setting to exploit the hierarchical relationship between the scores. Similar to above, we first learn jointly the lower-level tasks (Cogency, Effectiveness, Reasonableness) 
resulting in three scores $\hat{y}_{\textnormal{Cog}}$, $\hat{y}_{\textnormal{Eff}}$ and $\hat{y}_{\textnormal{Rea}}$. Next, we employ these scores for informing the overall AQ predictor by concatenating these with the hidden document representation $\textbf{h}_\mathit{D}$:
$
    \mathbf{h}_{\mathit{informed}} = \mathbf{h}_{D}^\frown [\hat{y}_{\textnormal{Cog}}, \hat{y}_{\textnormal{Eff}},\hat{y}_{\textnormal{Rea}}].
$
The resulting vector $\mathbf{h}_{\mathit{informed}}$ serves as input to the overall AQ predictor as defined in before. 

\section{Experiments}
\label{sec:experiments}
We employ the proposed architectures above to answer research questions RQ2--RQ5.

\subsection{RQ2: Computational theory-based AQ assessment}
To test whether our corpus supports the development of theory-based AQ assessment models, this experiment employs all single-task models presented in Section~\ref{sec:methodology} (\textsc{Arg length}, \textsc{SVR$_{\textit{tfidf}}$}, \textsc{SVR$_{\textit{embd}}$} \textsc{Wachsmuth$_{\textit{CFS}}$}, and \textsc{BERT ST}).  We train and predict on the domain-specific data sets 
and report the results on the \emph{mix} test set per AQ dimension for each domain.\footnote{Trends for the other evaluation sets (crowd and expert) are similar. Full results can be found in the supplementary material.} 
Details on the hyperparameter optimization can be found in the appendix. 
%
%
%
%
%
\begin{table}[t]\centering
\small
\parbox{.46\linewidth}{
\setlength{\tabcolsep}{2pt}

\begin{tabular}{llccc}\toprule
&\textbf{Model} &\textbf{CQA} &\textbf{Debates} & \textbf{Reviews}  \\
\midrule
\multirow{7}{*}{\textbf{Overall}} &\textsc{Arg length} &0.406 &0.420  &0.365 \\
&\textsc{SVR$_{\textit{tfidf}}$} &0.389 &0.265  &0.450 \\
&\textsc{SVR$_{\textit{embd}}$} &0.278 &0.388  &0.265 \\
&\textsc{Wachsmuth$_{\textit{CFS}}$} &0.492  &0.432 &0.533 \\
&\textsc{BERT ST}  & \textbf{0.652} & \textbf{0.511} &\textbf{0.605} \\
\addlinespace
&\textsc{BERT MT$_{\textit{flat}}$} &\textbf{0.667} &\textbf{0.537} &0.588 \\
&\textsc{BERT MT$_{\textit{hier}}$} &0.661 &0.494 &0.593 \\
\midrule
\multirow{7}{*}{\textbf{Cogency}} & \textsc{Arg length}  &0.420  &0.437  &0.340 \\
&\textsc{SVR$_{\textit{tfidf}}$} &0.444 &0.257 &0.384 \\
&\textsc{SVR$_{\textit{embd}}$} &0.261 &0.333 &0.103 \\
&\textsc{Wachsmuth$_{\textit{CFS}}$} &0.503 &0.429  &0.464 \\
&\textsc{BERT ST} & \textbf{0.587} & \textbf{0.503} & \textbf{0.554} \\
\addlinespace
&\textsc{BERT MT$_{\textit{flat}}$} &0.633 &\textbf{0.541} &\textbf{0.561} \\
&\textsc{BERT MT$_{\textit{hier}}$}  &\textbf{0.638} &0.474 &0.541 \\
\bottomrule
\end{tabular}
}
\parbox{.53\linewidth}{
\begin{tabular}{llccc}\toprule
&\textbf{Model} &\textbf{CQA} &\textbf{Debates} & \textbf{Reviews}  \\
\midrule
\multirow{7}{*}{\textbf{Effectiveness}} & \textsc{Arg length} &0.390  &0.399 &0.372 \\ 
&\textsc{SVR$_{\textit{tfidf}}$} &0.411 &0.120 &0.340 \\
&\textsc{SVR$_{\textit{embd}}$} &0.293 &0.403 &0.187 \\
&\textsc{Wachsmuth$_{\textit{CFS}}$} &0.523 &0.450 &0.432 \\
&\textsc{BERT ST} & \textbf{0.612} & \textbf{0.542} &\textbf{0.555} \\
\addlinespace
&\textsc{BERT MT$_{\textit{flat}}$}  &\textbf{0.671}  &\textbf{0.570}  &0.514 \\
&\textsc{BERT MT$_{\textit{hier}}$}  &0.670 &0.532 &0.486 \\
\midrule
\multirow{7}{*}{\textbf{Reasonableness}} & \textsc{Arg length}  &0.396  &0.377  &0.405 \\
&\textsc{SVR$_{\textit{tfidf}}$} &0.457  &0.247  &0.452 \\
&\textsc{SVR$_{\textit{embd}}$}&0.379 &0.258 &0.234 \\
&\textsc{Wachsmuth$_{\textit{CFS}}$} &0.476 &0.399 &0.432 \\
&\textsc{BERT ST} &\textbf{0.665} & \textbf{0.418} & \textbf{0.609} \\
\addlinespace
&\textsc{BERT MT$_{\textit{flat}}$} &0.644 &\textbf{0.473} &0.610 \\
&\textsc{BERT MT$_{\textit{hier}}$}  &0.626 &0.408 &\textbf{0.611} \\
\bottomrule
\end{tabular}
}
\caption{Pearson correlations of our model predictions with the annotation scores on the mix test annotations when training on in-domain data. Numbers in bold indicate best performances.
}\label{tbl:in-domains-1}\label{tbl:in-domains}
\end{table}
\paragraph{Results.}
The respective Pearson correlation scores for AQ dimensions on the three domain-specific test sets are shown in Table~\ref{tbl:in-domains-1}.
Generally, we reach medium to high Pearson correlation scores of up to nearly $0.67$. 
However, like the IAA, performance varies across domains: 
On Debates, the best model, \textsc{BERT ST}, 
achieves a correlation coefficient with the annotation scores for reasonableness of $0.42$ 
and on the CQA forums, it achieves a performance of $0.67$. The BERT-based regressor 
outperforms the other methods, showing that we can successfully utilize a large-scale corpus with theory-based AQ dimensions to train models for automatic AQ assessment (\textbf{RQ2}). 
Note that \textsc{Arg Length} is relatively high across all domains and properties and often outperforms \textsc{SVR$_{\textit{tfidf}}$} and \textsc{SVR$_{\textit{embd}}$}, indicating a slight length bias in the corpus. 
However, \textsc{BERT ST} outperforms this baseline in all cases by a large margin, demonstrating this model's ability to capture useful information beyond pure length. 
%
%
%
%
%
%
%
%
%
%
%
%
%
%

\subsection{RQ3: Effect of AQ dimension interrelations}
\label{sec:mtlearning}
Next we seek to determine whether it is possible to exploit the interrelations between the three dimensions and the overall AQ by conducting experiments on \corpus.  
We compare the multi-task learning architectures, \textsc{BERT MT$_{\textit{flat}}$} and \textsc{BERT MT$_{\textit{hier}}$}, against the results of the \textsc{BERT ST} model, the best performing single-task model. Again, we train and predict on the domain-specific data splits. 
%
%
%
%
%

\paragraph{Results.}
Table~\ref{tbl:in-domains} shows the respective Pearson correlation scores for the four AQ dimensions on each domain.
%
%
 The multi-task learning models outperform the single-task model in $9$ out of $12$ cases,
  which suggests that the interrelations between the AQ dimensions and overall AQ can be exploited to improve model performance (\textbf{RQ3}). More specifically, the best method is \textsc{BERT MT$_{\textit{flat}}$}, which outperforms the other methods in $7$ cases.  \textsc{BERT ST} and \textsc{BERT MT$_{\textit{hier}}$} are best in $3$ and $2$ cases, respectively. 

\subsection{RQ4: Unified multi-domain model}
\label{sec:multidomain}
We examine whether our corpus supports training a unified multi-domain model.
To this end, we train the BERT-based models on the joint training set covering all domains and test performance on each individual domain, thereby including out-of-domain data in training. Similarly, we optimize the hyperparameters on the joint development set. We compare with the best in-domain score from Table~\ref{tbl:in-domains}.

\paragraph{Results.} The respective results for the four argument quality dimensions 
can be seen in Table~\ref{tbl:all-domains}. 
\begin{table}[!t]\centering
\parbox[t]{.48\linewidth}{
\strut\vspace*{-\baselineskip}\newline
\setlength{\tabcolsep}{2pt}
\small
\begin{tabular}{llrrr}\toprule
&\textbf{Model} &\textbf{CQA} & \textbf{Debates} & \textbf{Reviews} \\\midrule
\multirow{4}{*}{\textbf{Overall}} &Best in-domain  &0.667  &0.537 &0.605 \\
&\textsc{BERT ST} &0.676 &0.545  &0.596 \\
&\textsc{BERT MT$_{\textit{flat}}$}  &\textbf{0.681} &\textbf{0.562}  &\textbf{0.633} \\
&\textsc{BERT MT$_{\textit{hier}}$}  &0.665 &\textbf{0.562}  &0.622 \\
\midrule
\multirow{4}{*}{\textbf{Cogency}} &Best in-domain &0.638  &0.541  &0.561 \\
&\textsc{BERT ST}  &0.608 &0.515 &0.563 \\
&\textsc{BERT MT$_{\textit{flat}}$} & \textbf{0.653} &0.542 &0.570 \\
&\textsc{BERT MT$_{\textit{hier}}$} &0.638 &\textbf{0.552}  &\textbf{0.599} \\
\midrule
\multirow{4}{*}{\textbf{Effectiveness}} &Best in-domain &0.671  &0.570  &0.555 \\
&\textsc{BERT ST} &\textbf{0.686}  &\textbf{0.598} &0.601 \\
&\textsc{BERT MT$_{\textit{flat}}$}  &0.670  &0.578 &\textbf{0.603} \\
&\textsc{BERT MT$_{\textit{hier}}$} &0.653 &0.592 &0.576 \\
\midrule
\multirow{4}{*}{\textbf{Reasonableness}} &Best in-domain  &\textbf{0.665}  &0.473  &0.611 \\
&\textsc{BERT ST}  &0.635  &\textbf{0.487} &0.603 \\
&\textsc{BERT MT$_{\textit{flat}}$}  &0.657  &0.486  &0.631 \\
&\textsc{BERT MT$_{\textit{hier}}$}  &0.633 &0.483 &\textbf{0.643} \\
\bottomrule
\end{tabular}
\caption{Pearson correlations of the model predictions with the annotation scores when training on the joint training sets of all domains. We compare with the best result of the in-domain setting.}\label{tbl:all-domains} 
%
}
\hfill
\parbox[t]{.48\linewidth}{
\strut\vspace*{-\baselineskip}\newline
\centering
\small
\begin{tabular}{llcc}\toprule
\textbf{Domain} &\textbf{Dimension} &\textbf{$r$} &\textbf{$\rho$} \\\midrule
\textsc{BERT IBM} & -- & 0.492 &	0.456 \\
\newcite{gretz2019large} & -- & 0.52 & 0.48 \\
\midrule
All &Overall &\textbf{0.313} &\textbf{0.303} \\
&Cogency &0.311 &0.300 \\
&Effectiveness &\textbf{0.313} &\textbf{0.303} \\
&Reasonableness &0.304 &0.298  \\
\midrule
CQA Forums &Overall &0.258 &0.224 \\
&Cogency &\textbf{0.269} &\textbf{0.228}  \\
&Effectiveness &0.262 &0.225 \\
&Reasonableness &0.262 &0.226  \\
\midrule
Debate Forums &Overall &\textbf{0.336} &\textbf{0.326} \\
&Cogency &0.331 &0.321 \\
&Effectiveness &\textbf{0.336} &\textbf{0.326}  \\
&Reasonableness &0.333 &0.319  \\
\midrule
Review Forums &Overall &0.150 &0.145  \\
&Cogency &0.139 &0.138 \\
&Effectiveness &\textbf{0.152} &\textbf{0.151} \\
&Reasonableness &0.149 &0.148  \\
\bottomrule
\end{tabular}
\caption{Performance of \textsc{BERT MT$_{\textit{flat}}$} trained on \corpus, predicting on IBM-Rank-30k evaluated against the weighted average score. 
}\label{tbl:ibm-predict-results}
}
\end{table}
In $11$ out of $12$ cases, training on all domains increases the performance compared to the best in-domain model. While the models are less domain-specific, the increased amount of data leads to better convergence and lead to gains up to $5$ percentage points. 

\subsection{RQ5: Synergies between practical and theory-driven AQ}
To empirically test the hypothesis that synergies exist between practical and theory-based AQ assessment, we conduct a bi-directional experiment with the recently released IBM-Rank-30k \cite{gretz2019large}.

\paragraph{Experimental setup.} IBM-Rank-30k consists of 30,497 crowd-sourced arguments relating to 71 topics, where each argument is restricted to 35--210 characters. The corpus has binary judgments indicating whether raters would recommend the argument to a friend. Based on these ratings, a score for each argument was computed, either using MACE or weighted average of all ratings.  
 Compared to \corpus, IBM-Rank-30k is much larger but the arguments are much shorter and more artificial than real world texts. Manual inspection revealed that the nature of the texts substantially differs from each those in \corpus, i.e., arguments mainly cover reasons for higher-level claims. For example, in IBM-Rank-30k for the topic \emph{``We should end racial profiling"}, a highly rated argument  is \emph{``racial profiling unfairly targets minorities and the poor"}. 

We conduct three experiments in two directions: (E1) train on \corpus, then predict on IBM-Rank-30k, (E2) train on IBM-Rank-30k, then predict on \corpus, and finally,  (E3) train on IBM-Rank-30k, next, train on \corpus, and then, predict on \corpus.

For \textbf{experiment (E1)}, we take the (already trained) \textsc{BERT MT$_{\textit{flat}}$} models trained on each domain of \corpus and predict on the test portion of IBM-Rank-30k. This enables us to determine which one of our domains and dimensions are closest to the data and annotations in IBM-Rank-30k. 
We compare against the best score reported in the \newcite{gretz2019large} as well as against our own reimplementation using BERT$_{\textit{BASE}}$, dubbed \textsc{BERT IBM}.\footnote{Note that \newcite{gretz2019large} do not indicate whether they employ BERT$_{\textit{BASE}}$ or BERT$_{\textit{LARGE}}$.} 
We optimize the \textsc{BERT IBM} baseline by grid searching for the learning rate $\lambda \in \{2e-5, 3e-5\}$ and the number of training epochs $\in \{3,4\}$ on the IBM-Rank-30k development set.
For the already trained models from Sections~\ref{sec:mtlearning} and \ref{sec:multidomain}, no further optimization is necessary. 
In \textbf{experiment (E2)}, we reverse the direction of (E1): We train a BERT-based regressor as defined before 
on the MACE-P aggregated annotations of IBM-Rank-30k.\footnote{This corresponds to our \textsc{BERT IBM} baseline from before.} 
We predict on \corpus and correlate the scores with our annotations. 
Finally for \textbf{experiment (E3)}, in order to flatten out expected losses from the zero-shot domain transfer, inspired by \newcite{phang_stilts} we use IBM-Rank-30k in the Supplementary Training on Intermediate Labeled Tasks-setup (STILT), i.e., we take the trained \textsc{BERT IBM} encoder and continue training the model as \textsc{BERT IBM MT$_{\textit{flat}}$} in the all-domain setup.
We compare both models from (2) and (3) with the \textsc{BERT MT$_{\textit{flat}}$}.

\paragraph{Results.} 
\begin{table}[t]\centering
\small
\parbox{.46\linewidth}{
\setlength{\tabcolsep}{2pt}

\begin{tabular}{llrrr}\toprule
& &\textbf{CQA} &\textbf{Debates} &\textbf{Reviews} \\\midrule
\multirow{3}{*}{\textbf{Overall}} &\textsc{BERT IBM}  &0.392  &0.317  &0.154 \\
&\textsc{BERT IBM MT$_{\textit{flat}}$}  &0.666 &0.543 &0.568 \\
&\textsc{BERT MT$_{\textit{flat}}$} &\textbf{0.681}  &\textbf{0.562}  &\textbf{0.633} \\
\midrule
\multirow{3}{*}{\textbf{Cogency}} &\textsc{BERT IBM} &0.368  &0.274  &0.149 \\
&\textsc{BERT IBM MT$_{\textit{flat}}$}  &0.639 &0.518 &0.541 \\
&\textsc{BERT MT$_{\textit{flat}}$} &\textbf{0.653}  &\textbf{0.542} &\textbf{0.570} \\
\bottomrule
\end{tabular}
}
\parbox{.53\linewidth}{
\begin{tabular}{llrrr}\toprule
& &\textbf{CQA} &\textbf{Debates} &\textbf{Reviews} \\\midrule
\multirow{3}{*}{\textbf{Effectiveness}} &\textsc{BERT IBM} &0.426 &0.378  &0.195 \\
&\textsc{BERT IBM MT$_{\textit{flat}}$}  &\textbf{0.678}  &\textbf{0.594}  &0.545 \\
&\textsc{BERT MT$_{\textit{flat}}$}  &0.670 & 0.578  &\textbf{0.603} \\
\midrule
\multirow{3}{*}{\textbf{Reasonableness}\hspace{-2pt}} &\textsc{BERT IBM} &0.348  &0.246  &0.151 \\
&\textsc{BERT IBM MT$_{\textit{flat}}$}  &0.637  &0.465  &0.581 \\
&\textsc{BERT MT$_{\textit{flat}}$} &\textbf{0.657}  &\textbf{0.486}  &\textbf{0.631} \\
\bottomrule
\end{tabular}
}
\caption{Pearson correlations on \corpus when predicting with \textsc{BERT IBM} (trained on IBM-Rank-30k) and \textsc{BERT IBM MT$_{\textit{flat}}$} trained on IBM-Rank-30k in STILT setup fine-tuned on \corpus in comparison to \textsc{BERT MT$_{\textit{flat}}$}.}\label{tbl:stilt}
\end{table}

%
%
%
%
%
%
%
%
%
%
In experiment (E1) (Table~\ref{tbl:ibm-predict-results}), as expected, the zero-shot domain transfer results in a large drop compared to training on the associated train set of IBM-Rank-30k. Quite surprisingly, the model trained on the debate forums reaches the highest correlation scores -- even higher than the model trained on \emph{all-domains}. Further, in most cases, the effectiveness predictions correlate best with the annotations provided by \newcite{gretz2019large}. This is in-line with the authors' observations, who manually had to annotate the data for the theory-based scores.
%
%

Table~\ref{tbl:stilt} displays the results of (E2)--(E3). Experiment (E2), draws a similar picture: zero-shot domain transfer using \textsc{BERT IBM} results in a huge loss in performance compared to \textsc{BERT MT$_{\textit{flat}}$}. 
Finally, the results in (E3) indicate potential for using resources drawn from practical approaches in a theory-based AQ assessment scenario: When reusing the encoder in the STILT setup, \textsc{BERT IBM MT$_{\textit{flat}}$}, the losses originating from the zero-shot domain transfer can be flattened out -- in some cases even outperforming \textsc{BERT MT$_{\textit{flat}}$}. This is especially the case when correlating the predictions with our annotations for the effectiveness dimensions. 
To sum up, our experiment (E1)--(E3) yield the following findings: (1) Large-scale predictions, obtained from a theory-based AQ model on a large (practical) AQ data set, correlate mostly with the Effectiveness dimension.
(2) The transferred knowledge obtained in the STILT-setup on IBM-Rank-30k in \textsc{BERT IBM MT$_{\textit{flat}}$} improves the performance on \corpus for Effectiveness the most. 
These two facts match \newcite{gretz2019large}'s hypothesis  
that their annotations mostly captured Effectiveness. 
We empirically substantiate the idea (without any manual effort) that a theory-based approach can inform practical AQ research and increase interpretability of practically-driven research outcomes and, on the other hand, the practical approach can increase the efficacy of theory-based AQ models when targeting a certain domain and dimension. 

\section{Conclusion and Future Work}
\label{sec:conclusion}
 Specific assessment of the rhetorical, logical, and dialectical perspectives on argumentative texts can inform researchers, e.g., about phenomena captured with annotations, 
 and help people improve their writing skills. 
 However, the field of computational AQ assessment has been almost exclusively driven by practical approaches. 
 Aiming to fill this gap, in this work we advance theory-based computational AQ research with the following contributions:
 
We performed a large-scale annotation study on English argumentative texts covering debate forums, Q\&A forums, and business review forums. We thereby presented \corpus, the largest and first multi-domain corpus annotated with theory-based AQ scores (\textbf{RQ1}).\footnote{Available from \url{https://github.com/grammarly/gaqcorpus} with annotation guidelines and interface.}
 Next, we proposed the first computational theory-based AQ models (\textbf{RQ2}) and demonstrated that jointly predicting AQ scores can improve the performance of the models (\textbf{RQ3}) and that in most cases, models benefit from including out-of-domain training data 
 (\textbf{RQ4}).
Finally, we investigated concrete synergies between the practical and the theory-based approach to AQ assessment in a bi-directional experimental setup (\textbf{RQ5}). The theory-based models can help to increase the interpretability of practical approaches, and practical approaches can be employed to increase performance of the theory-based models. 
In the future, we would like to deploy the models and study to what extent users can actually improve their argumentative writing by getting theory-based AQ feedback. Further, we will seek to develop ways of adding even finer-grained aspect scores at scale; this remains an open problem. 



\section*{Acknowledgements}
The work of Anne Lauscher is supported by the Eliteprogramm of the Baden-Württemberg Stiftung (AGREE grant).
We thank our linguistic expert annotators for providing interesting insights and discussions as well as the anonymous reviewers for their helpful comments. We also thank Henning Wachsmuth for consulting us w.r.t. his previous work and Yahoo! for granting us access to their data.

\bibliographystyle{coling}
\bibliography{coling2020}

\clearpage

\end{document}